%% file: main.tex
\pdfoutput=1
\documentclass[11pt]{article}

\usepackage[preprint]{acl}

\usepackage{times}
\usepackage{latexsym}

\usepackage[T1]{fontenc}
\usepackage[utf8]{inputenc}
\usepackage{multirow}

\usepackage{microtype}
\usepackage{booktabs} % For professional looking tables
\usepackage{siunitx}  % Allows for alignment by decimal point

\usepackage{inconsolata}
\usepackage{graphicx}
\usepackage{xspace}
\usepackage{hyperref}
\usepackage{cleveref}
\usepackage{enumitem}
\usepackage[normalem]{ulem}
\crefformat{section}{\S#2#1#3}
\crefformat{subsection}{\S#2#1#3}
\crefformat{subsubsection}{\S#2#1#3}
\crefrangeformat{section}{\S#3#1#4 to~\S#5#2#6}
\crefmultiformat{section}{\S#2#1#3}{ and~\S#2#1#3}{, #2#1#3}{ and~#2#1#3}
\Crefformat{figure}{#2Fig.~#1#3}
\Crefmultiformat{figure}{Figs.~#2#1#3}{ and~#2#1#3}{, #2#1#3}{ and~#2#1#3}
\Crefformat{table}{#2Tab.~#1#3}
\Crefmultiformat{table}{Tabs.~#2#1#3}{ and~#2#1#3}{, #2#1#3}{ and~#2#1#3}
\Crefformat{appendix}{#2Appx.~\S#1#3}
\crefformat{algorithm}{Alg.~#2#1#3}
\Crefformat{equation}{#2Eq.~#1#3}
\newcommand{\modelname}{\textsc{FamiCom}\xspace}
\newcommand{\fullname}{\texttt{\underline{Fami}}larity and \texttt{\underline{Com}}plexity Based Performance Estimation\xspace}
\newcommand{\stitle}[1]{\vspace{1ex} \noindent{\bf #1}}

\title{\includegraphics[scale=0.15]{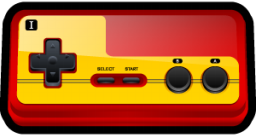}\modelname: Further Demystifying Prompts for Language Models with Task-Agnostic Performance Estimation}

\author{Bangzheng Li$^{1}$ \quad Ben Zhou$^{2}$ \quad Xingyu Fu$^{2}$ \quad Fei Wang$^{3}$ \quad Dan Roth$^{2}$ \quad Muhao Chen$^{1}$ \\
    $^{1}$University of California, Davis \quad
  $^{2}$University of Pennsylvania \\
  $^{3}$University of Southern California \\
  \texttt{bzhli@ucdavis.edu} 
}

\begin{document}
\maketitle
\input{content/0_abstract}

\input{content/1_introduction}

\input{content/2_method}
\input{content/3_validation}

\input{content/4_promptSelection}
\input{content/5_indirectICL}
\input{content/6_relatedWork}
\input{content/7_conclusion}

\section*{Ethical Considerations}
Innovations in technology often encounter the moral challenge of dual-use: the same development can bring both benefits and risks. With the probing method and benchmark presented in this paper, the line between beneficial and harmful usage largely depends on data. Proper utilization of the technology necessitates the legal and ethical acquisition of input text corpora and other modalities of inputs. Legal frameworks and standards are crucial for ensuring proper data use and for granting individuals the right to remove their data. In the absence of such regulation, the ethical use of data depends on the responsibility of technology users. Additionally, the generated and analysis data may exhibit biases that systematically affect accuracy for less represented groups or in new areas, potentially resulting in performance disparities among sub-populations based on ethnicity, race, gender, and other factors. Moreover, the effectiveness of trained systems diminishes when applied to new data that deviates from their training set. Therefore, issues of generalizability and fairness must be thoroughly examined when implementing the methodologies discussed in this paper. It is crucial to embed ethical considerations as fundamental principles at each stage of system development, ensure high levels of transparency and clarity in data, algorithms, models, and functions within the system, release software under open-source licenses to facilitate public scrutiny and investigate strategies to safeguard at-risk groups.

\section*{Limitations}
Our work proposes that he inherent perceived complexity of the end tasks should be included in the estimation of prompt performance, together with the familiarity of LMs to the prompt. To this end, we identify the following limitations. 

\stitle{Limited validation experiments} We only conducted the validation experiments on Mistral-7B as the massive data volume to process in this experiment. With more effort in the future, we can extend to more families of large language models.

\stitle{Limited demonstrations}
The current Indirect ICL is still a pilot study and does not include a large scale of demonstrations? Future works may benefit from wider range of tasks involved in the indirect ICL experiments and bring more insights in the prompting studies.

\iffalse
\section*{Acknowledgments}

This document has been adapted
by Steven Bethard, Ryan Cotterell and Rui Yan
from the instructions for earlier ACL and NAACL proceedings, including those for
ACL 2019 by Douwe Kiela and Ivan Vuli\'{c},
NAACL 2019 by Stephanie Lukin and Alla Roskovskaya,
ACL 2018 by Shay Cohen, Kevin Gimpel, and Wei Lu,
NAACL 2018 by Margaret Mitchell and Stephanie Lukin,
Bib\TeX{} suggestions for (NA)ACL 2017/2018 from Jason Eisner,
ACL 2017 by Dan Gildea and Min-Yen Kan,
NAACL 2017 by Margaret Mitchell,
ACL 2012 by Maggie Li and Michael White,
ACL 2010 by Jing-Shin Chang and Philipp Koehn,
ACL 2008 by Johanna D. Moore, Simone Teufel, James Allan, and Sadaoki Furui,
ACL 2005 by Hwee Tou Ng and Kemal Oflazer,
ACL 2002 by Eugene Charniak and Dekang Lin,
and earlier ACL and EACL formats written by several people, including
John Chen, Henry S. Thompson and Donald Walker.
Additional elements were taken from the formatting instructions of the \emph{International Joint Conference on Artificial Intelligence} and the \emph{Conference on Computer Vision and Pattern Recognition}.
\fi

\bibliography{anthology,custom}
% \bibliography{custom}

\appendix

% \section{Example Appendix}
% \label{sec:appendix}

% This is an appendix.

\end{document}

%% file: content/0_abstract.tex
\begin{abstract}
Language models have shown impressive in-context-learning capabilities, which allow them to benefit from input prompts and perform better on downstream end tasks. Existing works investigate the mechanisms behind this observation, and propose label-agnostic prompt metrics that can better estimate end-task performances. One popular approach is using perplexity as a way to measure models' familiarity with the prompt. While showing consistent improvements on in-domain tasks, we found that familiarity metrics such as perplexity cannot accurately estimate performance in complicated situations such as task or domain transferring scenarios. 
In this work, we propose a revised measure called \modelname{}, providing a more comprehensive measure for task-agnostic performance estimation. 
Specifically, \modelname{} combines familiarity with \textit{complexity} -- the inherent difficulty of end tasks, which is an important factor missing from current metrics. 
Experiments show that \modelname{} strongly correlates with end-task performances, producing a 0.85 Spearman's correlation, versus 0.43 of familiarity-only ones'. We further apply \modelname{} to automatic prompt and demonstration selection, and outperform existing methods and baselines by more than \textbf{7.0\%} in accuracy.

\end{abstract}

% \begin{abstract}
% Language models have shown impressive in-context-learning capabilities, which allow them to benefit from input prompts and perform better on downstream end tasks. Existing work has been investigating the mechanisms behind this observation by finding label-agnostic prompt metrics that correlate with or ``project'' end-task performances. One popular approach is using perplexity as a way to measure models' familiarity with the prompt, which does not work well in task and domain transferring scenarios. In this work, we argue that these metrics are missing an important factor, which we call complexity: the inherent perceived difficulty of the end tasks. We propose a revised measure called \modelname{}, which combines the factors of familiarity and complexity. In our experiments, we show that \modelname{} strongly correlates with end-task performances, producing a 0.85 Spearman's correlation, versus using familiarity-only's 0.43. As a result, we can apply \modelname{} as a way for automatic prompt and demonstration selection, outperforming existing methods and baselines by more than 7\%.
% \end{abstract}

%% file: content/1_introduction.tex
% \vspace{1mm}
\section{Introduction}

\begin{figure*}[t!]
\begin{center}
    \includegraphics[width=1\textwidth]{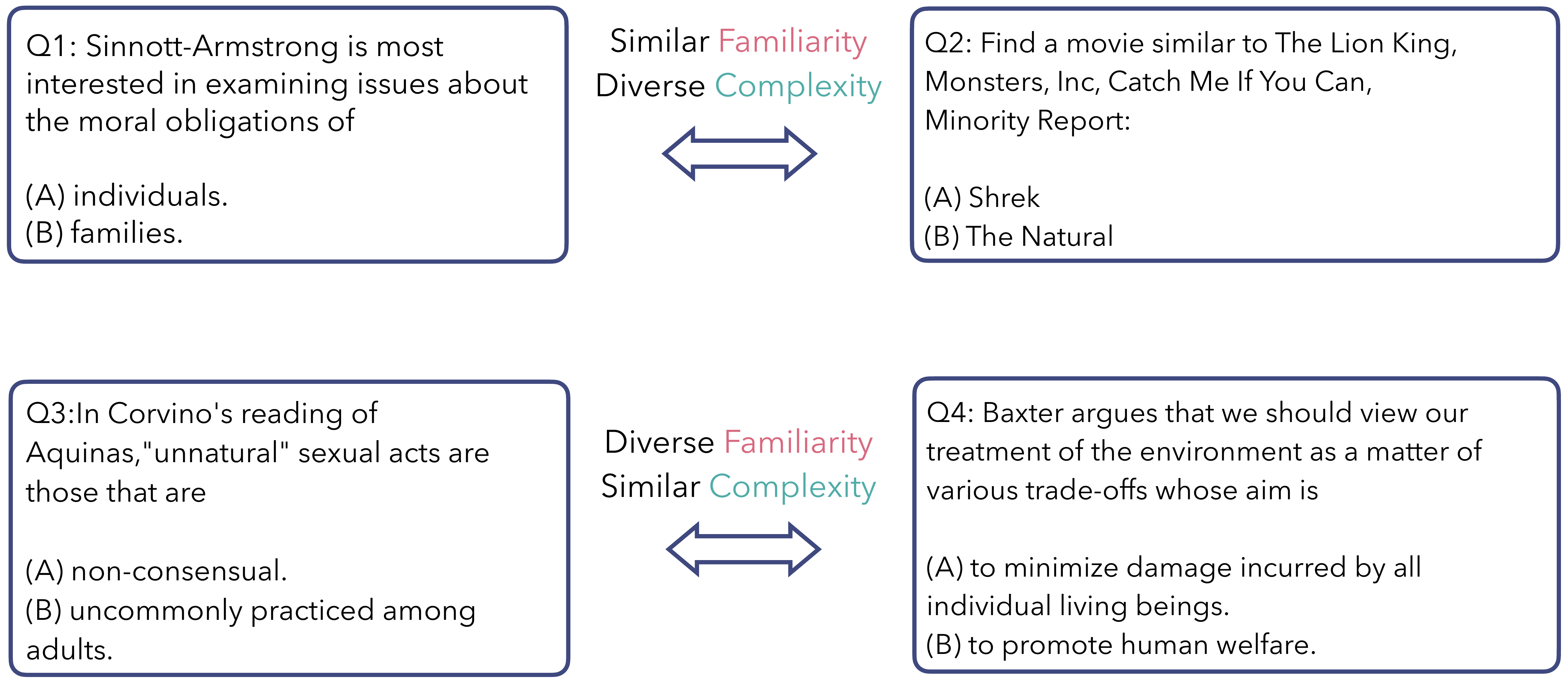}
    %\vspace{-2em}
    \caption{Prompts can have similar familiarity yet diverse complexity, or diverse familiarity but similar complexity. Estimating LM's performance merely on single factor is not enough.}
    \label{fig:example}
    \vspace{-1em}
\end{center}
\end{figure*}

% \muhao{at least put a couple citations in the first paragraphs}
Recent works have shown that large language models (LLMs) can perform new NLP tasks by following simple instructions or seeing only a few in-context examples \cite{rubin-etal-2022-learning,NEURIPS2020_1457c0d6,liu-etal-2022-makes,wei2022chain}. 
% They demonstrated impressive performances on various tasks by prompting with detailed instructions, reasoning processes or exemplary prompt-answer pairs, rather than task-specific fine-tuning. However, it is still challenging to investigate \textit{why} textual prompts can contribute to the end-task performances and \textit{what} factors would affect the effectiveness of prompts or demonstrations.
Even though these models have demonstrated impressive capabilities to efficiently use the information in the input prompt to improve downstream task-specific performances, we still do not fully understand \textit{why} textual prompts can contribute to end-task performances. This research direction is crucial because it will contribute to better automatic prompt design and few-shot example selection methods, in addition to helping us peek into the internal mechanisms of LLMs.

% \muhao{explain that we think how difficult the prompt is w.r.t. the LM is another important factor first, then term it as complexity.}
% \muhao{need to tell first why we need the }
% Earlier works proposed to assess how familiar a model is to the prompt by the perplexity of the sequence~\cite{gonen-etal-2023-demystifying,m-etal-2023-ctqscorer} and revealed the perplexity-accuracy correlation of the model. It is intuitive to estimate the model's expected performance indirectly by its \textbf{familiarity} to the prompts. However, our experiments also suggest that such correlation diminishes in more complex reasoning tasks \Cref{sec:res}. The accuracy-perplexity trend observed on classification tasks are no longer obvious on tasks with more reasoning steps. Therefore we propose to introduce another factor to better assess the model performance under different textual inputs, which enables a series of downstream application such as prompt selection or in-context-learning example selection, which enables assessing the quality of the prompt, selecting or demystifying prompts without the need of knowing any tasks beforehand.

One promising way to investigate the relations between input prompts and end-task performances is to propose label-agnostic measures regarding the input prompts. If a measure strongly correlates with the end-task performance without seeing the task labels, it would suggest that such a measure is in the right direction to understand why the input prompt helps. Earlier works have proposed to use perplexity as a measure in this direction, which evaluates how familiar a model is to the prompt~\cite{gonen-etal-2023-demystifying,m-etal-2023-ctqscorer}. While it is intuitive to estimate the model's expected performance indirectly by its \textbf{familiarity} to the prompts, we have found that such correlation diminishes in more complex reasoning tasks \Cref{sec:validation}. The correlation between perplexity and accuracy is no longer obvious on tasks with more reasoning steps. Therefore, we propose introducing another factor better to assess the model performance under different textual inputs.

\Cref{fig:example} showcases examples where prompts to the LM can have similar familiarity yet diverse complexity, or diverse familiarity but similar complexity. Considering Q1 and Q2, both of which have nearly same values in model familiarity\footnote{Mistral-7B, familiarity\textsubscript{sim} defined in \Cref{ssec:fam}.} but are intrinsically different in complexity. Q1 asks about the work of Walter Sinnott-Armstrong, a prominent philosopher who has extensively explored interpersonal relationships. With these knowledge stored in parameters, a LM can easily gives the correct answer of (B) families. However, Q2 asks about which movie is similar to the given five movies. It requires the model to first extract the features of each movie, find the common, and compare with each options. This process involves more reasoning steps than Q1, and questions like Q2 is therefore have lower expected accuracy. Q3 and Q4 demonstrated a different situation. They are both reading comprehension questions from high school textbooks. To answer them, model can simply retrieve the parametric knowledge and select the option closest to the word distribution in training data. The reasoning steps are similar which yield close complexity. However, they have different familiarities which may results in different expectation of accuracy.
% We hypothesize that the end-task performances of LLMs have a strong correlation with not only the \textbf{familiarity} of the model to the query prompt but also the \textbf{complexity} of the prompt. To be specific, %given %any format of textual input to the model, various ways to prompt 
% if we consider different ways to prompt the LLM to address the same task,
% %the more familiar the model is to the input and the less complexity the question is, the better performance the model on the input will be.
% a better prediction performance by the LLM is more likely achieved with a more familar and less complex prompt.

Motivated by such observations, we argue that the familiarity factor alone are not robust enough across tasks. An important factor is missing, which we hypothesize to be the complexity of the end task. In other words, familiarity correlations only hold for tasks that have relatively similar complexity levels, and we should factor complexity in if we want to build a label-agnostic metric that will transfer across domains and tasks. To be specific, if we consider different ways to prompt the LLM to address the same task, a better prediction performance by the LLM is more likely achieved with a more familiar and less complex prompt.

To demonstrate such intuition with quantitative analyses, we first investigate several strategies to measure the intrinsic \textbf{complexity} of any given prompt for the LLM. In addition to zero-shot or few-shot prompting the model's own complexity assessment on the task prompt, we also device a practical technique to translating a given task prompt to a Transformer programming language \cite{zhou2023algorithms} and assess its operation-level complexity.
Inspired by \citet{gonen-etal-2023-demystifying}, we involved perplexity into the quantification of model familiarity. Moreover, as the recent study \cite{eureqa} suggests that the mutual similarities among key words in a reasoning question also have a noticeable correlation with model performance, we combined both of the perplexity and similarity to redefine the model's familiarity to the query question.

We conducted a cross-task prompting evaluation: Given a input question from a multiple choice task, we evaluate the performance of models with cross-task demonstration. The selection of evaluation tasks span across 28 different multiple choice question answering tasks yielding exponential demonstration-input question combinations which is enough to come to a statistical conclusion. We show empirically that the proposed task-agnostic measure, namely \modelname (\fullname), has a positive correlation with model performance across a diverse set of tasks and models which provide a insight of how prompting affect the model performance. To further demonstrate the application of our %factor
measure, we devise a method on prompt selection task. Our results revealed that \modelname has a better guidance than similarity search or perplexity ranking which proved the effectiveness of our method.

To further demonstrate the potential impact of \modelname, we propose a novel \emph{indirect in-context learning} setting where given any end task, the model is allowed to retrieve any available annotated demonstrations that are not originally labeled for the end task. Our experimental results also suggest that \modelname consistently outperforms similarity search and provide better demonstration benefit to the model,
allowing for more effective retrieval of indirect supervision \cite{yin2023indirectly} of in-context demonstrations from the wild.
% As task-specific demonstration is not always available in a real-world application of LLMs, we suggest this would also

To summarize, the contribution of our work is three-fold. First, we propose a task-agnostic prompt performance estimation measure \modelname describing the relationship between model performance, model familiarity to a prompt and the complexity of the input question. Second, we prove the effectiveness of our formula by evaluating it with a massive cross-task prompting experiment. Finally, to elaborate the potential application of our formula, we tested it on prompt selection and an innovative indirect ICL task.

%% file: content/2_method.tex
%\section{What makes some prompts different than others?}
\section{What measures a good prompt?}
% \ben{I believe metric is a better word choice than measure, but your call.}

In this section, we introduce \modelname -- a task-agnostic measure for estimating the effectiveness of textual prompt for LLMs. The calculation of \modelname is based on the hypothesis that \textbf{the lower complexity of the prompt and higher familiarity of the model with regard to the prompt correlates with better performance}. We start with the definition and calculation of two key factors of \modelname: \textbf{complexity}(\Cref{ssec:complex}) and \textbf{familiarity}(\Cref{ssec:fam}). We also discuss other controllable parameters and give the concluded calculation of \modelname in \Cref{ssec:factor}.

\subsection{Complexity Estimation}\label{ssec:complex}
As language models scale up and become more and more capable, benchmark questions and tasks also tend to become more ``complex'' to effectively evaluate the capabilities and limitations of these models. These evaluations have evolved from ``simple'' tasks such as sentiment analysis~\cite{socher-etal-2013-recursive}, natural language inference~\cite{williams-etal-2018-broad}, to more ``complex'' multi-task understanding~\cite{hendrycks2020measuring} and open-domain question answering~\cite{rajpurkar-etal-2016-squad}. The testing questions evolves from determining whether a comment is positive to answering high-school biology questions. 

However, defining the \textbf{complexity} of %these tasks
task prompts remains challenging due to the inherently ambiguous nature of natural language prompts. In this work, we make some presumptions for the approximation of prompt complexity. First, the complexity is language model based. In other words, for fixed prompts of a task, the measured difficulty may vary according to the choice of language model. Second, the complexity of a question %is defined as the number of steps it takes to solve the problem
is proportional to how many steps (or sub-problems) it needs to solve the problem \cite{khot2022decomposed}. The atomic steps in this presumption may differ according to the first presumption. We propose three methods to approximate the prompt complexity, namely \texttt{direct complexity}, \texttt{guided complexity} and \texttt{operational complexity}. 

The \textbf{direct complexity} is measured by querying the model with a simple prompt $p_{complex}$: \textit{``How many steps does it takes to solve the problem.''}. Formally, for an input prompt $q$, the complexity of the prompt for language model $L$ is defined as $L(p,q)$. 

The \textbf{guided complexity} is similar to the \textbf{direct complexity} with extra guidance of human written demonstrations. We add detailed human written examples to the prompt $p_{complex}$, in which each example contains a sample question, a list of steps of the question and the final result for the steps. The examples involves questions of different steps to have a broader guidance.

The \textbf{operational complexity} is based on the pseudo programming language of RASP-L \cite{zhou2023algorithms}. It is a human-readable programming language which defines programs that can be compiled into Transformer weights. Each atomic operation in RASP-L is a calculation in a Transformers block. In this way, the steps are equivalent to %the line of code
the quantity of operations needed to write a RASP-L program to answer the prompt $q$. One key challenge of this measurement is that each prompt needs a individually composed RASP-L program which makes the method impossible to scale up. To address such issue, we append human written %(prompt, RASP-L program) pairs 
demonstrations containing other prompts and their RASP-L programs to the evaluated prompt $p_{complex}$ and query the language model to compose the RASP-L program and calculate the complexity. 

For each prompt, we can query the model multiple (such as $k=5$) times to get the complexity score and compute the average as the final complexity for the prompt.

\subsection{Familiarity Estimation}\label{ssec:fam}

We consider two ways of estimating the familiarity of a prompt to the language model.

The concept of \textbf{familiarity} is intended to approximate how a model is familiar with the input text. \citet{gonen-etal-2023-demystifying} proposed to use the perplexity of a prompt as a proxy for its occurrences in the data. It is based on the intuition that ``a sequence that is more expected by the model is more likely to aid the model to extract the relevant information.''. The experimental results showed a positive correlation between model performances and the perplexity of the prompt. We adopt perplexity as the method of \textbf{familiarity\textsubscript{ppl}} measurement.

In addition, \citet{eureqa} also found that the similarities among key words in a question, such as key entities, have a correlation with model accuracy in chained reasoning tasks. This inspired of our second method to measure familiarity. 
Here we focus on the ``key'' tokens and their mutual similarities. We first extract the ``key'' tokens by selecting Top-K tokens that have the highest perplexities in a prompt. 
Inspired by the counterfactual theories \cite{lewis2013counterfactuals} and prior studies on probing token salience \cite{kaushik2019learning,wang-etal-2021-table-based,qi-etal-2021-onion},
we adopt the intuition the intuition that any word that leads to high entropy given the previous contexts are likely important words such as preposition or pronounce. Those words with lower probability can be conveying the important information of the prompt. Among the selected tokens, we compute the mutual cosine similarities of the embedding of the tokens in the input text and use the averaged similarities as \textbf{familiarity\textsubscript{sim}}.

\subsection{Combining Familiarity and Complexity}\label{ssec:factor}
Based on the previous analysis, the measured factors of \modelname can be incorporated in the polynomial:
% \muhao{Caution: *-b*. globally make sure b is non-negative}
\begin{equation}
    \modelname = f^a \cdot c^{-b} \\
\end{equation}
where $f,c$ are familiarity and complexity respectively. $a,b$ are positive hyper-parameters that we will investigate in \Cref{sec:validation} 
where we also show that both properly chosen $a,b$ values 
that has a positive correlation with model performance.

%% file: content/3_validation.tex
\section{Validation Analysis}\label{sec:validation}

This section presents the analytical experiment for validating the \modelname measure. We first introduce the task settings and the data used for analysis in \Cref{ssec:validTaskInfo}, and subsequently elaborate the model configurations and evaluation protocol in \Cref{ssec:validTaskConfig}\footnote{The configurations discussed here also apply to experiments in \Cref{sec:promptSelect} and \Cref{sec:indiICL}}. Finally, \Cref{ssec:validTaskAnalysis} presents the experimental results and the analysis.

\subsection{Task Description}\label{ssec:validTaskInfo}
Since \modelname is a task-agnostic performance estimation, we design a \textbf{cross-task prompting} scenario for the validation. Given a prompt from a specific task, we randomly pair it with the demonstrations of other tasks to evaluate the model performance. The \modelname measure is computed for each of the input. 
We conduct cross-task prompting evaluation like this in a scale to expose the statistical relationship between the measure and the true model performance.

The 28 evaluation task pool are sampled from MMLU, BigBench together with StrategyQA and CommonsenseQA. For each of these tasks, we retrieve its Chain-of-Thought prompts from the CoT Hub~\cite{fu2023chainofthought}, which is an open-source project for measuring LLMs' reasoning capabilities with resourceful prompts for various benchmarks. %Each task 
During inference, a question is paired with three randomly selected CoT demonstrations for other tasks in the pool. We randomly sample 200 instances for each task, along with three chain-of-thought demonstrations per task. This preparation method yields approximately 100,000 question-demonstration pairings for our experiment, thereby guaranteeing its statistical significance. 

Every sampled task is multiple choice question answering\footnote{StrategyQA has a ``Yes or No'' question type, which can be easily formatted as ``Options: (A) Yes (B) No''} and we follow the prompt format in the CoT Hub where each question is given as ``[Question] Options:[list of options]''. Each option is labeled with an alphabetical letter, for example ``(A) Jupiter'', which help locating the choice from the generative LMs' response.

It is noteworthy that these questions differ in the number of options which may introduce biases in evaluation, especially when computing the complexity. A question with more options may be regarded as more complex by the model. To alleviate this bias, each of the sampled questions is reduced to a binary option question. To do so, the correct option is retained in the two options where the other is a randomly picked false choice.  The  sequence of options is shuffled to mitigate the potential biases. 

\input{table/corr}
\begin{figure}[t!]
    \includegraphics[scale=0.36]{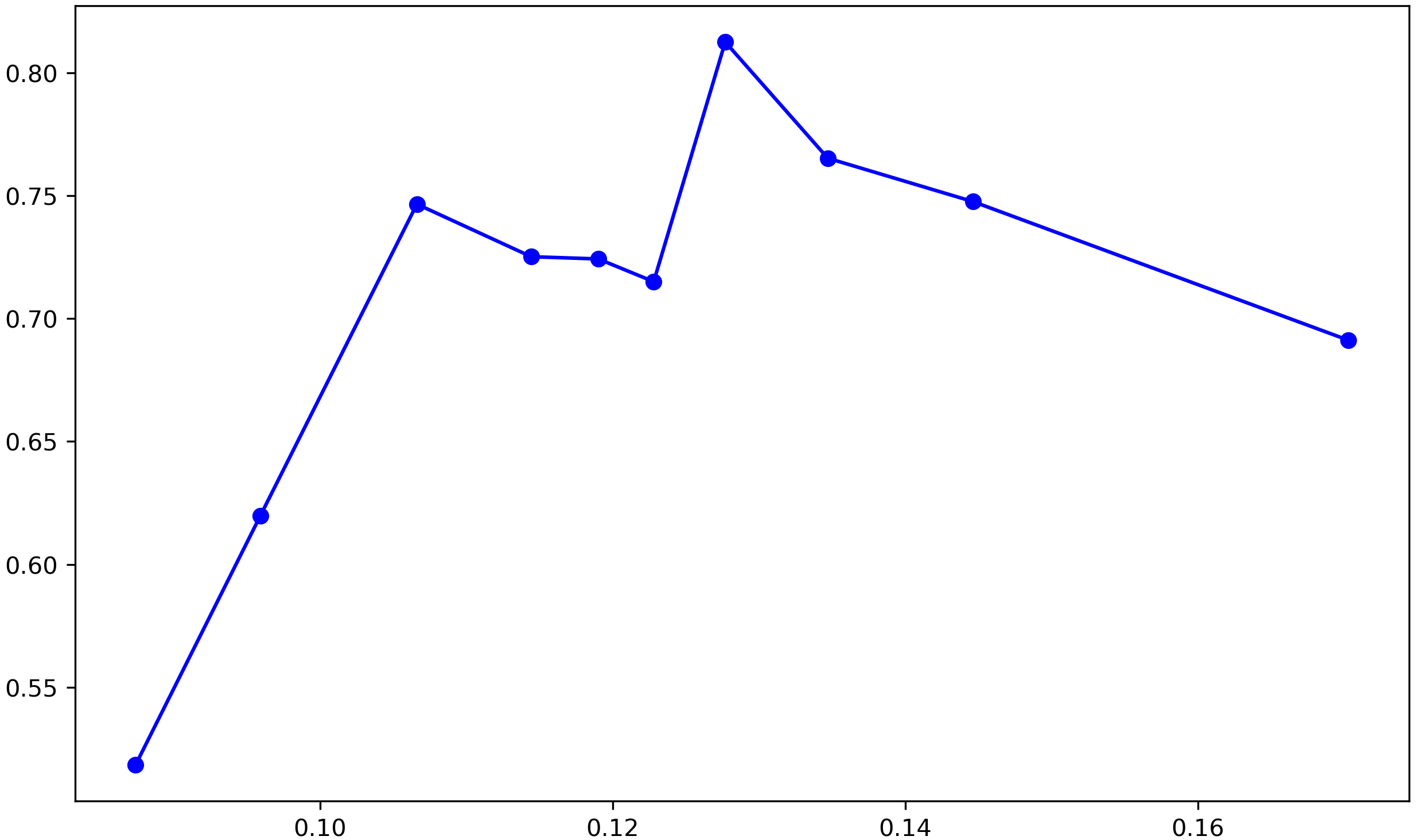}
    \caption{The correlation between Mistral performance on validation experiment and input \textbf{familiarity}. Familiarity itself doe not show a clear correlation with performance.}
    \label{fig:fam}
\end{figure}

\begin{figure}[t!]
    \includegraphics[scale=0.36]{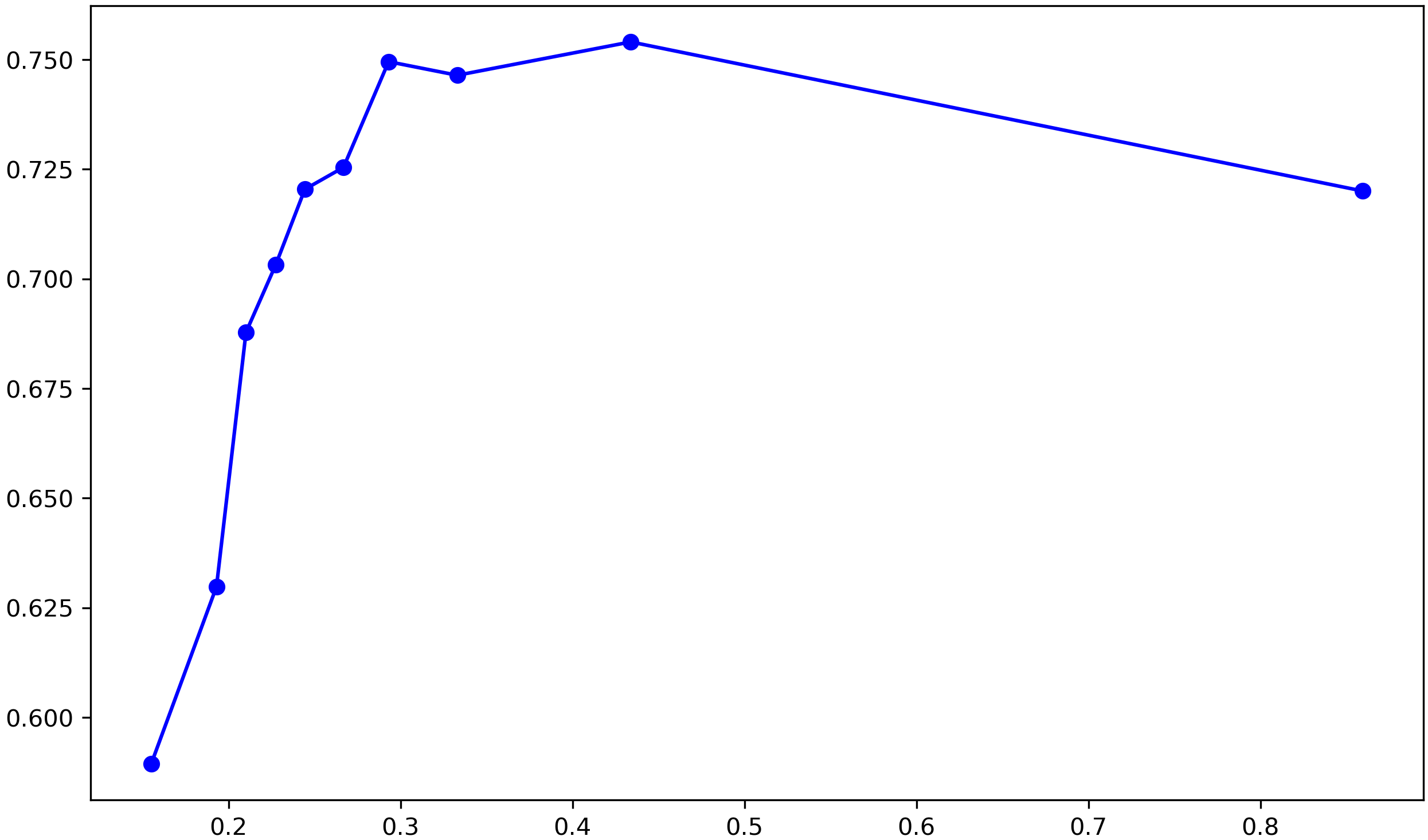}
    \caption{The correlation between Mistral performance on validation experiment and the input's inverse \textbf{complexity}. Complexity demonstrates a better correlation with performance but needs further calibration.}
    \label{fig:com}
\end{figure}

\begin{figure}[t!]
    \includegraphics[scale=0.36]{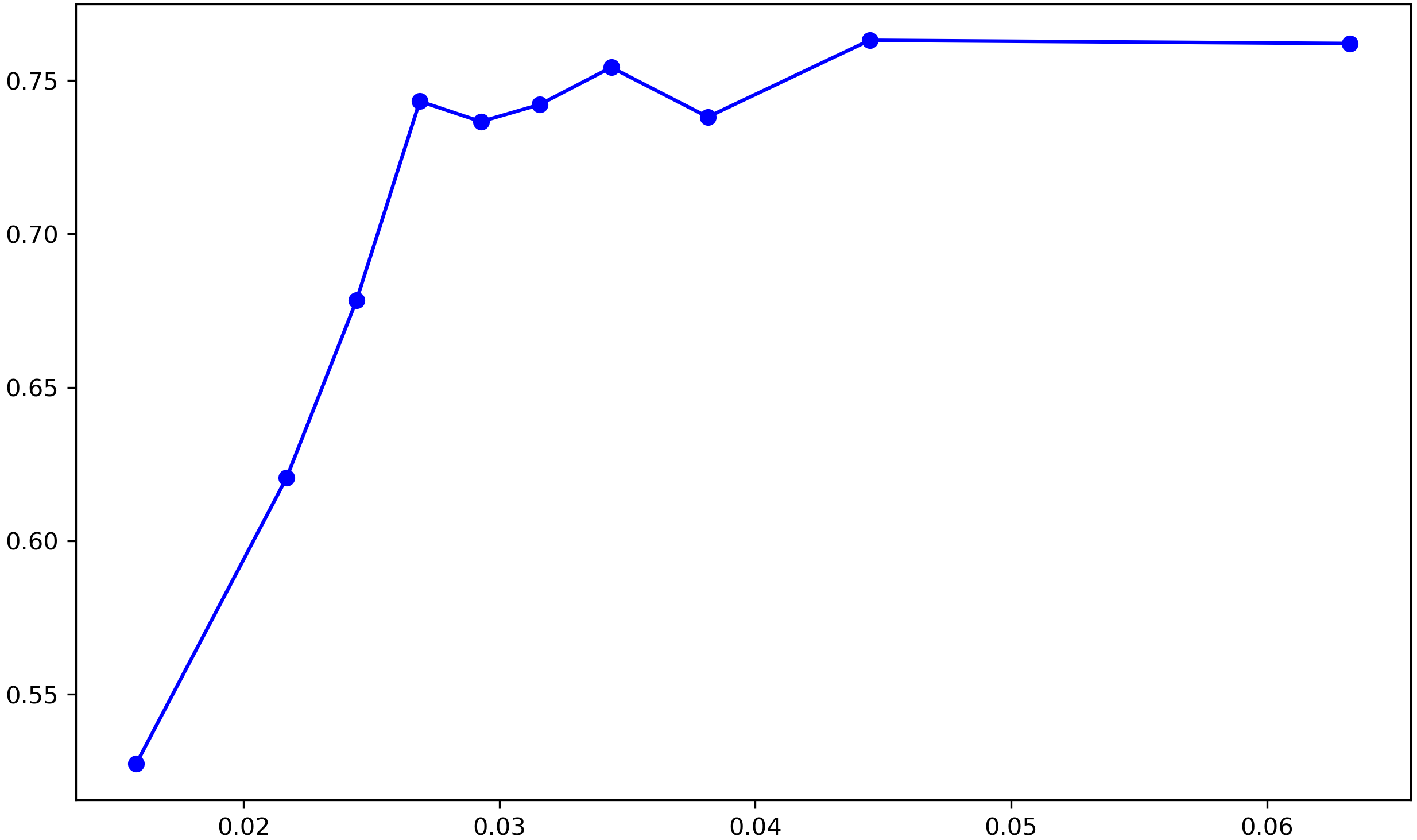}
    \caption{The correlation between Mistral performance on validation experiment and the input \textbf{\modelname} measure. Compared to \Cref{fig:fam,fig:com}, the trend line is much more consistent, and the improvement is more steady.}
    \label{fig:famcom}
\end{figure}

\subsection{Model configuration and Evaluation Protocol}\label{ssec:validTaskConfig}
We evaluate on open-source LLMs in different sizes due to the fact that this scale of evaluation is too costly on closed-source LLMs and the token distributions or perplexities are not accessible in their APIs. The tested models include Phi-3-mini-128k, Mistral-7B-instruct-v0.2, and Llama-2-13b-chat. All models run with a temperature $\tau = 0.8$. Experiments are run on a machine with 8*Nvidia ADA 6000 GPUs.

We evaluate the \textbf{accuracy} of LLMs towards the correct answer. Since all the prepared data are multiple-choice QA with labeled options, a response is considered correct if it contains the correct label. We also applied the self-consistency strategy with majority voting in five runs for all experiments following ~\cite{wang2022self}.

\subsection{Validation Analysis Results} \label{ssec:validTaskAnalysis}
We show that our hypothesis hold and \modelname measure has a correlation with the model performance. \Cref{fig:fam}, \Cref{fig:com}, \Cref{fig:famcom} provide visualizations between the model performance and familiarity, complexity and \modelname measures respectively. These figures adopts \textbf{familiarity\textsubscript{sim}} and the \textbf{guided complexity} which are tested to have better correlation. The token similarities are computed over tokens with top 20 perplexities. The coefficients are fixed as $f^1 \cdot c^{-1}$ after the hyper-parameter search. \Cref{tab:corr} depicts the Spearman correlation of model performance against these measures.

It is shown in \Cref{fig:fam} that \textbf{familiarity} itself does not highly correlates with the model performance. This conclusion can also be drawn from its Spearman Correlation $\rho = 0.426 < 0.5$ with $p=0.002$. The \textbf{complexity} factor, on the other hand, has a positive correlation with model performance with $\rho=0.695$ and $p<<0.00625$\footnote{Bonferroni Test following \citet{gonen-etal-2023-demystifying}}. This meets the intuition that model perform worse on more complex questions and also validate the effectiveness of our approximation of \textbf{complexity}. The \modelname measure has a better monotonic correlation with model performances, whose Spearman correlation value reaches 0.848 with $p<<0.00625$. It can then be concluded that our hypothesis hold in a large scale of experiments and \modelname can effectively estimate the performance of prompts on LLMs.

%% file: table/corr.tex
% \begin{table}[h]
% \centering
% \begin{tabular}{|c|c|c|}
% \toprule
% \textbf{Measure} & \textbf{Spearman corr.} & \textbf{\(p\) value} \\
% \midrule
% \textbf{Familiarity} & 0.426 & 0.002 \\
% \midrule
% \textbf{Complexity} & 0.695 & 2e-8 \\
% \midrule
% \textbf{ModelName} & 0.848 & 7e-9 \\
% \bottomrule
% \end{tabular}
% \caption{Spearman Correlation of model performances and proposed factors or measures.}
% \label{tab:trans_corr}
% \end{table}

\begin{table}[t]
\centering
\scalebox{1}{{\fontsize{11pt}{17pt}\selectfont
\begin{tabular}{|l|c|c|}
\hline
\textbf{Measure} & \textbf{Spearman Corr.} & \textbf{P-value} \\ \hline
Familiarity & 0.426 & 0.002 \\ \hline
Complexity & 0.695 & 2e-8 \\ \hline
\textbf{\modelname} & 0.848 & 7e-9 \\ \hline
\end{tabular}
}}
\caption{Enhanced Table of Spearman Correlations and P-values for Model Performance against Proposed Factors.
As we can see, \modelname surpasses single usage of familiarity or complexity.
}
\label{tab:corr}
\end{table}

%% file: content/4_promptSelection.tex
\section{The Prompt Selection Task}\label{sec:promptSelect}
\input{table/promptSelect}
In this section, we discuss a prompt selection task to further explore the effectiveness and application of \modelname. \Cref{ssec:promptSelectTask} introduces the task, \Cref{ssec:promptSelectData} discusses the evaluation datasets and the following \Cref{ssec:promptSelectRes} analyzes the experimental results. In this task, we only examine the Mistral model\footnote{Following \citet{liu2024monotonic}} and using the same set of evaluation metrics as \Cref{ssec:validTaskConfig}.

\subsection{Task Description}\label{ssec:promptSelectTask}
To inspect if \modelname can properly estimate model performance on different prompts as observed, we conduct evaluation on \textsc{Super-NaturalInstructions}(\textsc{Sup-NatInst} for short, \citet{wang-etal-2022-super}). This benchmark contains expert-contributed instructional prompts for 1,616 NLP tasks for evaluating the zeo-shot performance of LLMs. In \textsc{Sup-NatInst}, every task is provided with a instruction that includes the task's definition for transforming an input text into a specified output, along with multiple examples to illustrate both the desired and the undesired results. We only use the task definition as the instructional prompt and use GPT-4 to generate four more task descriptions in \textsc{Sup-NatInst} style for each task involved.

To calculate the \modelname measure for each candidate prompt, we pair it with the query question and compute their combined \textbf{familiarity}. The salient words in the question can therefore be combined with the salient words in the prompt in the calculation.

\subsection{Datasets}\label{ssec:promptSelectData}
We choose five discriminative tasks tested in \cite{gonen-etal-2023-demystifying} from Huggingface Dataset\footnote{https://huggingface.co/docs/datasets/index} for a broad evaluation. These includes: (i)\textbf{AG News} \cite{zhang2015character} for news topic classification (ii) \textbf{IMDB} \cite{maas-etal-2011-learning} for sentiment analysis on movie reviews (iii) \textbf{GLUE-Cola} \cite{warstadt-etal-2019-neural} for grammatical acceptability discrimination; (iv) \textbf{Emotion} \cite{saravia-etal-2018-carer} for emotion classification on tweets; (v) \textbf{Tweet Offensive} \cite{barbieri-etal-2020-tweeteval} for offensive tweet discrimination. Each dataset is sampled 1,000 test data a balanced evaluation of prompt selection. 

\subsection{Baselines}
We compare our method with two baseline: \textbf{SPELL} \cite{gonen-etal-2023-demystifying} and the original prompt in \textsc{Sup-NatInst}. \textbf{SPELL}(\textbf{S}electing \textbf{P}rompts by \textbf{E}stimating \textbf{L}M \textbf{L}ikelihood) selects the prompts with the lowest perplexity for a given task after manually creating a set of candidate prompts and expanding them to hundred-scale using automatic paraphrasing and back-translation. The original(Ori) \textsc{Sup-NatInst} prompt is expert-created for each task.

\subsection{Prompt Selection Results}\label{ssec:promptSelectRes}
\Cref{tab:promptSelect} shows the performances of different method. \modelname consistently achieves the best performance on most of the tasks except for Cola. Specifically, \modelname improves the accuracy of Mistral-7B by 4.6\% on average across all tasks, surpassing the improvement of 3.3\% offered by SPELL. This suggests that involving both \textbf{familiarity} and \textbf{complexity} can give a better estimation of prompt performance than the single \textbf{familiarity} factor.

%% file: table/promptSelect.tex
\begin{table*}[h]
\centering
\setlength{\tabcolsep}{15pt}
\scalebox{1}{{\fontsize{10pt}{17pt}\selectfont
\begin{tabular}{c|c|c|c|c|c}
\toprule
 Method & AG News & Imdb & Cola & Emotion & Offensive \\
 \midrule
Ori & 67.7 & 84.1 & 39.8 & 44.3 & 67.0 \\ 
 \midrule
SPELL & 78.0 & 86.0 & \textbf{43.4} & 46.8 & 65.1 \\
 \midrule
\modelname & \textbf{78.9} & \textbf{87.6} & 41.0 & \textbf{51.4} & \textbf{67.1} \\
\bottomrule
\end{tabular}
}}
\caption{Prompt selection results given in accuracy. The best result for each task is in \textbf{bold}. As we can see, \modelname consistently performs better than Ori and SPELL on almost all of the datasets.}
\label{tab:promptSelect}
\end{table*}

%% file: content/5_indirectICL.tex
\input{table/expSelect}
\section{Indirect In-context Learning}\label{sec:indiICL}

In this section, we demonstrate a novel Indirect In-context Learning (ICL) task to further showcases the practicality of \modelname. We start by explaining the task ( \Cref{ssec:indiICLTask}), discuss the data preparation (\Cref{ssec:indiICLData}), and demonstrate the baseline(\Cref{ssec:indiICLBaseline}). The model configurations follow \Cref{ssec:validTaskConfig} and the results are shown in \Cref{ssec:indiICLRes}.

\subsection{Task Description}\label{ssec:indiICLTask}
\vspace{2mm}
The term ``indirect'' in the name suggests that the focus of this task setting is not on task-specific demonstration selection. 
Instead, indirect ICL considers a generalized pool of examples from different tasks together, and the goal is to identify most contributive examples for each specific input. 
%This setting generalizes the regular ICL to leverage any available labeled examples that are not necessarily dedicated to the end-task, including those that may provide helpful indirect supervision \cite{yin-etal-2023-indirectly}.
Each example comprises a question and its corresponding chain-of-thought response. Indirect ICL generalizes regular ICL to allow for any available labeled examples that are not necessarily dedicated to the end-task, including those that may provide helpful incidental supervision \cite{yin-etal-2023-indirectly} to aid in-context learning. 
This task also addresses a practical situation where searching for the most beneficial demonstration when the user query or the end task is not known beforehand.

\vspace{2mm}
\subsection{Data Preparation}\label{ssec:indiICLData}
\vspace{2mm}
In this task, we collect examples from each of 28 tasks in a pool. Each task contributes three examples and each example consists of a question and its chain-of-thought response. The pool finally gathers 84 CoT examples for later inference. We evaluate the model on 17 different tasks from MMLU, BigBench, and StrategyQA. For each task, 200 examples at most are randomly sampled for evaluation. We retrieved the CoT demonstration from CoT Hub~\cite{fu2023chainofthought} and formatted the CoT examples as well as the testing questions in the same way described in \Cref{ssec:validTaskInfo}.

\subsection{Baselines}\label{ssec:indiICLBaseline}
We compare \modelname with K-Nearest Neighbors(KNN), which ranks and selects the candidate examples based on their similarities with the input prompt. To be specific, we extract the embeddings of the last tokens of an example and the input prompt, calculate the cosine similarity between them as the distance. Same as \modelname, we select top $K=3,5,7$ in this experiment for a broader inspection.

\subsection{Indirect ICL Results}\label{ssec:indiICLRes}
\Cref{tab:expSelect} provides detailed experimental results for Indirect ICL. \modelname outperforms KNN on the majority of tasks across various K values. Specifically, the macro-average score of \modelname surpasses that of KNN by 7.1\%, 3.8\%, and 2.1\% for K=3, 5, and 7, respectively. The improvement is more pronounced as the number of examples decreases, indicating that \modelname is particularly beneficial in few-shot settings. In general, the results for Indirect ICL also proves the advantage of involving complexity approximation into similarity-based familiarity for prompt performance estimation.

%% file: table/expSelect.tex
\begin{table*}[h]
\centering
% \small
\scalebox{1}{{\fontsize{9pt}{16pt}\selectfont
\begin{tabular}{c|c|c|c|c|c|c|c}
\hline
&\#Examples & \multicolumn{2}{c|}{K=3} & \multicolumn{2}{c|}{K=5} & \multicolumn{2}{c}{K=7} \\ 
% \cmidrule(r){1}\cmidrule(r){2-3} \cmidrule(lr){4-5} 
\midrule
Source&Task&\modelname &KNN&\modelname &KNN&\modelname &KNN\\
\midrule
\multirow{11}{*}{\rotatebox[origin=c]{90}{MMLU}} & medical-genetics	&	82.6	&	76.8	&	79.7	&	69.6	&	82.6	&	79.7\\
&professional-psychology	&	78.3	&	63.6	&	76.9	&	71.3	&	79.7	&	73.4\\
&formal-logic	&	66.7	&	54.9	&	71.6	&	62.7	&	68.6	&	63.7\\
&moral-disputes	&	81.9	&	70.8	&	81.9	&	77.8	&	83.3	&	75.7\\
&public-relations	&	78.8	&	71.8	&	76.5	&	76.5	&	81.2	&	70.6\\
&computer-security	&	85.9	&	76.9	&	79.5	&	82.1	&	83.3	&	82.1\\
&astronomy	&	87.2	&	68.8	&	88.1	&	78.0	&	85.3	&	83.5\\
&abstract-algebra	&	63.7	&	50.0	&	63.7	&	50.0	&	65.0	&	57.5\\
&nutrition	&	85.2	&	81.2	&	87.2	&	85.9	&	89.3	&	89.9\\
&high-school-biology	&	82.9	&	76.0	&	85.6	&	80.8	&	82.9	&	88.4\\
&business-ethics	&	76.7	&	75.3	&	79.5	&	78.1	&	71.2	&	78.1\\
\midrule
\multirow{1}{*}{\rotatebox[origin=c]{0}{ }}& StrategyQA	&	57.0	&	56.0	&	62.5	&	58.5	&	60.5	&	55.5\\
\midrule
\multirow{5}{*}{\rotatebox[origin=c]{90}{BIG}} &tracking-shuffled-objects-seven-objects	&	52.5	&	54.5	&	47.5	&	56.0	&	47.5	&	57.0\\
&formal-fallacies	&	25.5	&	16.5	&	30.5	&	14.5	&	29.0	&	17.5\\
&hyperbaton	&	70.0	&	69.5	&	75.0	&	74.5	&	76.5	&	74.0\\
&tracking-shuffled-objects-three-objects	&	47.0	&	45.5	&	50.0	&	50.5	&	48.5	&	54.5\\
&logical-deduction-five-objects	&	83.0	&	77.0	&	78.5	&	83.0	&	80.5	&	78.0\\
\midrule
\multirow{1}{*}{\rotatebox[origin=c]{0}{ }}&Macro-Avg	&	70.9	&	63.8	&	71.4	&	67.6	&	71.5	&	69.4\\
\bottomrule
\end{tabular}
}}
\caption{Indirect ICL Results of \modelname and KNN with K=3,5,7 examples. \modelname outperforms KNN on the majority of tasks across various K values, indicating that \modelname outperforms similarity-based familiarity for prompt performance estimation.}
\label{tab:expSelect}
\end{table*}

%% file: content/6_relatedWork.tex
\vspace{2mm}
\section{Related Work}
\vspace{2mm}

\stitle{Zero-shot prompt tuning.}
There are numerous studies trying to refine the prompt for better outcomes from language models in recent years. They can be roughly classified in two categories: prompt selection and prompt rewriting. These tasks focus on improving the end task performance by retrieving or creating prompts that will improve the zero-shot performance of LMs. Based on gradient-guided search, \citet{shin-etal-2020-autoprompt} leverages an automated method to project a prompt onto discrete phrases to improve masked LM performances. \citet{deng-etal-2022-rlprompt} applied reinforcement leaning to optimize discrete prompts. 

To adapt to general-purpose LLMs, where the user query may be unknown, researches have been conducted to exploit the intrinsic word distribution of LLMs obtained from the training data and propose familiarity-based prompt refinement methods \cite{gonen-etal-2023-demystifying,wang-etal-2023-readprompt,lee2024crafting}. These works are based on a phenomenon that lower perplexities of prompts are preferred by LMs to perform better across a wide range of tasks.

\vspace{2mm}
\stitle{Few-shot demonstration selection.} Furthermore, some researches focus on in-context-learning (ICL) example selection instead of a single prompt to improve model performance with few-shot learning \cite{iter-etal-2023-context,lee2024crafting, lu-etal-2022-fantastically}. Works have also been done to further investigate what are the driving factors that contributes to the different performances in ICL \cite{min-etal-2022-rethinking}, which is closer to the scope of this work.

While the last sections of this work lie in zero-shot and few-shot prompt tuning, the main target on this work is stepping forward to study the essence of prompting: the heuristics of word distributions in LMs are not enough to estimate the prompt performance, the complexity of the query shall also be considered. Our paper proposed a measure that will better describe the effectiveness of prompt, without any training effort.

\vspace{2mm}
\stitle{Analytical studies on model performance.} Several studies have examined how the distribution of training data affects model performance on specific tasks. \citet{gonen-etal-2023-demystifying} suggested utilizing perplexity to estimate the distribution of input queries within the training dataset. \citet{razeghi-etal-2022-impact} explored the capability of language models to process numerical tasks involving terms infrequently encountered during pre-training. They found that models perform better on instances where the terms are more common in the training data.

While these studies suggested using word frequency to link pre-training data with task performance, \modelname emphasized that the \textbf{complexity} of the input question is another contributor to the estimation of performance.

%% file: content/7_conclusion.tex
\vspace{2mm}
\section{Conclusion}
\vspace{2mm}
In this work, we propose \modelname{}, a label-agnostic metric that predicts whether an input prompt can lead to high downstream end-task performance or not, without the need to understand what the task is about. Different from previous methods~\cite{min-etal-2023-nonparametric,gonen-etal-2023-demystifying} that only consider using familiarity such as perplexity, \modelname{} considers two dimensions: familiarity and complexity, which are shown both intuitively and empirically to be essential when building such metrics and solving cross-domain tasks. On the one hand, \modelname{} provides insights into the internal mechanisms of the in-context-learning capabilities of large language models; on the other hand, it can be used as an automatic prompt selection method because it predicts whether a prompt can produce high end-task performance. Our work motivates future works on building more fine-grained metrics using familiarity and complexity and inspires works on LLM interpretability.